\documentclass{article}

\PassOptionsToPackage{numbers, compress}{natbib}

\usepackage[preprint]{neurips_2026}

\newcommand{\tool}[0]{FlashEvolve}

\usepackage[utf8]{inputenc} 
\usepackage[T1]{fontenc}    
\usepackage{hyperref}       
\usepackage{url}            
\usepackage{booktabs}       
\usepackage{amsfonts}       
\usepackage{nicefrac}       
\usepackage{microtype}      
\usepackage{xcolor}         
\usepackage{graphicx}
\usepackage{subcaption}
\usepackage[table]{xcolor}
\usepackage{multirow}
\usepackage{enumitem}

\definecolor{addcolor}{HTML}{C2410C}  

\title{\tool{}: Accelerating Agent Evolution with Asynchronous Stage Orchestration}

\author{%
    Zhengding Hu\textsuperscript{1},
    Mingge Lu\textsuperscript{1},
    Zhen Wang\textsuperscript{1},
    Jixuan Ruan\textsuperscript{1},
    Chang Chen\textsuperscript{1},
    Zaifeng Pan\textsuperscript{1},\\
    \textbf{Yue Guan\textsuperscript{1},
    Ruiyi Wang\textsuperscript{1},
    Zhongkai Yu\textsuperscript{1},
    Chao Zhang\textsuperscript{2},
    Yufei Ding\textsuperscript{1}}\\
    \textsuperscript{1}University of California, San Diego
    \qquad
    \textsuperscript{2}Georgia Institute of Technology
}

\begin{document}

\maketitle

\begin{abstract}
LLM-based evolution has emerged as a promising way to improve agents by
refining non-parametric artifacts, but its wall-clock cost remains a major
bottleneck. We identify that this cost comes from synchronized stage execution
and imbalance inside each LLM-heavy stage. We present FlashEvolve, an efficient framework
that replaces synchronized execution with asynchronous workers and queues, allowing different stages and steps to overlap. To handle data staleness introduced by asynchrony, FlashEvolve tracks artifact
versions and applies different policies to update, discard, or patch stale
artifacts. Unlike weight-space staleness in asynchronous RL, language-space staleness is inspectable and repairable: a stale artifact is not just delayed work, but readable evidence that the LLM can reflect on, revise, and turn into useful evolution signal. FlashEvolve further improves throughput and token efficiency with
speculative stage completion and adaptive workflow control. On GEPA workloads, FlashEvolve improves proposal throughput by $3.5\times$ on local vLLM and $4.9\times$ on API serving over synchronous GEPA. The same design also applies to ACE and Meta-Harness.
\end{abstract}

\section{Introduction}

A growing line of recent work enables LLM agents to evolve themselves.
Instead of updating model weights, these systems iteratively refine
the non-parametric components that govern their behavior, including system prompts~\cite{agrawal2025gepa, wang2023promptagent, xiao2025promptmii}, context and memory~\cite{zhang2025ace, ouyang2025reasoningbank, zhang2025memevolve}, harness
code~\cite{lou2026autoharness, lee2026metaharness} and generated
programs~\cite{novikov2025alphaevolve, lange2025shinkaevolve, assumpccao2025codeevolve}. This emerging paradigm of test-time self-evolution~\cite{gao2025survey}
fundamentally relaxes the access requirements of 
weight-space adaptation: it requires neither the labeled trajectories 
used by supervised fine-tuning nor the gradient updates required by 
reinforcement learning. By having an LLM reflect on full execution traces rather than 
optimize against scalar rewards, this paradigm draws a richer 
learning signal from each rollout: GEPA~\cite{agrawal2025gepa} outperforms GRPO with an average gain of 6\% across six reasoning benchmarks, while Meta-Harness~\cite{lee2026metaharness} automatically discovers agent harnesses that surpass the best hand-engineered baselines on different domain-specific benchmarks.

Despite its algorithmic appeal, agent evolution remains expensive in wall-clock execution time. Existing evolution algorithms pursue ``faster'' evolution by improving the quality of each
step through stronger reflection~\cite{agrawal2025gepa, zhang2025ace, yuksekgonul2024textgrad}, better artifact proposal and search~\cite{lange2025shinkaevolve, lu2026empirical}, or larger-batch updates~\cite{li2026combee}, thereby reducing the number of steps needed. However, fewer evolution steps do not necessarily translate into shorter wall-clock time. For example, on IFBench, a single GEPA evolution step already takes $\sim$2 minutes; Combee~\cite{li2026combee} parallelizes proposal generation, but further stretches each step to $\sim$2.8 minutes. Reaching a stable improvement requires more than 2 hours on an H100 GPU. This cost further grows with data scale, making evolution runs slow to tune and deploy in practice.

\begin{figure}[t]
    \centering

    \begin{subfigure}[t]{0.82\linewidth}
        \centering
        \includegraphics[width=\linewidth]{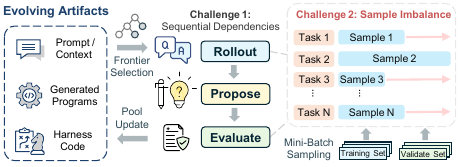}
        \label{fig:intro_left}
    \end{subfigure}

    \caption{Illustration of the multi-stage execution in agent evolution. The synchronized stage orchestration in existing implementations~\cite{agrawal2025gepa, zhang2025ace, li2026combee} exposes two efficiency challenges: sequential dependencies across stages and sample workload imbalance within each individual stage.}
    \label{fig:intro}
\end{figure}

Such high wall-clock cost comes from \textbf{synchronized stage execution}. As shown in Figure~\ref{fig:intro}, each evolution
step runs a sequence of LLM-heavy stages, such as running the current artifact on a mini-batch of inputs, proposing a new candidate artifact, and evaluating the new one. A later stage cannot start until the previous stage has fully completed. Such serial
structure prevents overlap across stages.

The cost inefficiency is amplified by \textbf{generation imbalance} inside each stage.
Request lengths vary widely across samples, such as different validation
samples in the evaluate stage. This creates a long-tail effect: the
longest requests determine the execution time of the whole stage. 
This reduces the effective batch size in both local serving frameworks~\cite{kwon2023vllm, zheng2024sglang} and
API-based remote calls, leading to low resource utilization and inefficient waiting for long samples.

To this end, we present \tool{}, a framework that improves the time efficiency of agent evolution through asynchronous stage orchestration. \tool{} treats
an evolution loop as a set of LLM-heavy stages connected by queues. This allows artifact
execution, proposal generation, evaluation, and pool update to overlap in time,
turning a synchronized loop into a streaming execution pipeline.

This design introduces new systems challenges. Asynchronous execution can
generate stale items because an artifact pool may change while earlier items
are still waiting in queues. \tool{} handles this with artifact-version
tracking and staleness-aware policies, including version comparison and discarding, or reflective patching for stale language artifacts. 
This property is specific to agent evolution. Unlike weight updates in SFT or reinforcement learning, evolution artifacts are prompts, memories, harness code, or programs. A stale artifact is therefore still an inspectable object: its relation to the current pool can be judged as complementary, redundant, or conflicting, and can be revised by the same LLM mechanism used for proposal. This makes staleness a semantic repair problem rather than only a scheduling hazard.
\tool{} further reduces waiting inside long stages through
speculative completion, and uses adaptive
workflow control to balance workload across stages. Together, these
mechanisms improve throughput while preserving the quality of evolution.

\section{Background and Motivation}
\label{sec:motivation}

\subsection{Agent Evolution: Self-Improvement Beyond Weight Updates}

Agent evolution has emerged as a new paradigm for adapting 
LLM-based systems to new data and tasks~\cite{gao2025survey, fang2025comprehensive}. This success stems from the already strong reasoning capability of modern LLMs~\cite{guo2025deepseekr1, jaech2024openaio1}, which enables a single model to reflect on its own trajectories~\cite{shinn2023reflexion}, critique its own outputs~\cite{madaan2023selfrefine}, and propose new artifacts that govern its own behavior, ranging from prompts, memory, and harness code that govern how the agent operates, to generated programs that constitute the task solution. Crucially, this happens without modifying model weights, 
sidestepping the training infrastructure of supervised fine-tuning~\cite{shoeybi2019megatron, zhao2023pytorchfsdp} and reinforcement learning~\cite{guo2025deepseekr1, ouyang2022rlhf} while delivering comparable or stronger 
gains.

For example, GEPA~\cite{agrawal2025gepa} and ACE~\cite{zhang2025ace} use reflection on execution traces to evolve system prompts and contextual playbooks. Meta-Harness~\cite{lee2026metaharness} and 
AutoHarness~\cite{lou2026autoharness} use a coding agent to evolve the harness based on prior runs and their failure modes. AlphaEvolve~\cite{novikov2025alphaevolve} and ShinkaEvolve~\cite{lange2025shinkaevolve} push this beyond the agent itself, evolving the generated programs the agent uses to solve problems, where the LLM acts as a mutation operator and an external evaluator scores each candidate.

An agent evolution loop iterates over multiple iteration steps, where each 
step consists of several stages, as illustrated in 
Figure~\ref{fig:intro}. The LLM-heavy stages are typically 
\emph{Generate}, \emph{Propose}, and \emph{Evaluate}. The 
Generate stage runs the current artifact on tasks to collect 
trajectories. The Propose stage reflects on these trajectories to 
produce a new candidate artifact. The Evaluate stage scores the 
candidate against task signals and filters out underperforming ones. A subsequent update commits the new artifact to the artifact 
pool. At the start of each step, new candidate artifacts are selected 
from the pool, through methods like Pareto-aware 
sampling~\cite{agrawal2025gepa} or evolutionary 
tournaments~\cite{novikov2025alphaevolve}.

\subsection{Inefficiency in Agent Evolution: Sequential and Imbalanced Stages}
\label{sec:inefficiency}

Despite its algorithmic appeal, agent evolution remains expensive in
wall-clock time. Based on our experiments, even with state-of-the-art LLM serving infrastructure such as vLLM~\cite{kwon2023vllm}, which supports continuous batching and prefix caching, GEPA with Qwen3-8B takes 50 minutes to complete 49 evolution steps on IFBench~\cite{pyatkin2025ifbench}, and 134 minutes to complete 411 steps on HotpotQA~\cite{yang2018hotpotqa}.

\begin{figure}
    \centering
    \includegraphics[width=\linewidth]{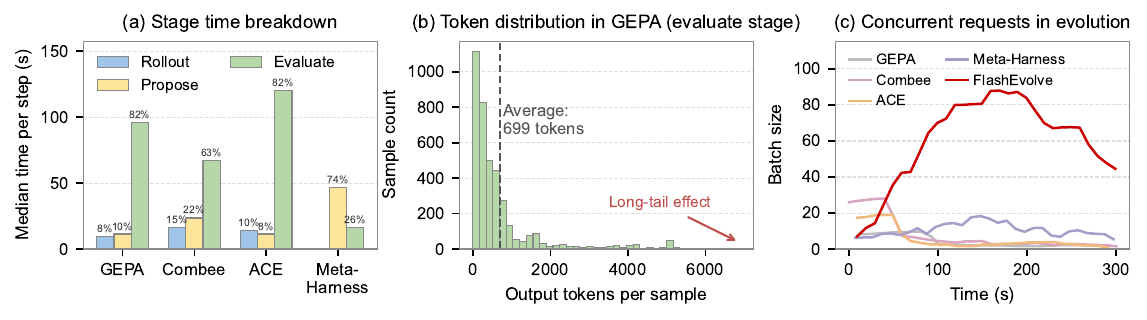}
    \caption{Profiling results of inefficiency in synchronized agent evolution. (a) Stage execution is serial, and stage time is highly imbalanced. (b) Within a single stage, output lengths show a long-tail distribution, so the slowest requests determine stage completion time.
(c) Serial stage execution and intra-stage imbalance reduce effective concurrency, while \tool{} keeps more requests in flight.}
    \label{fig:motivation}
\end{figure}

This inefficiency stems from sequential and synchronized stage execution. Each evolution step runs its LLM-heavy stages serially, and each stage internally waits for all parallel LLM requests to finish before advancing to the next stage. This structure produces two compounding costs. First, the serial chain forces total step time to be the sum of per-stage durations, with no opportunity to overlap stages. As shown in Figure~\ref{fig:motivation}(a), stage time is highly imbalanced, so different stages can become the bottleneck depending on the workload and algorithm. Second, the synchronization barrier at each stage's end forces the entire batch to wait for the slowest one. As shown in Figure~\ref{fig:motivation}(b), output lengths within a stage show a long-tail distribution, so a small number of long requests determine stage completion time. Consequently, sequential execution and intra-stage imbalance reduce effective concurrency and leave the LLM backend underutilized, as shown in Figure~\ref{fig:motivation}(c).

Such inefficiency cannot be solved by simply launching more LLM requests in parallel. Agent evolution must convert a synchronized multi-stage loop into a streaming workflow while preserving artifact-evolution semantics. This creates two challenges. First, asynchrony introduces artifact-level staleness: intermediate results may be produced from an artifact pool that has already changed before they are consumed. Second, naive parallel scaling can amplify workload imbalance: fast stages may overproduce items for slow stages, while long-tail requests within a stage can still delay downstream execution. This causes queue buildup, longer staleness windows, and wasted LLM work. These challenges require orchestration mechanisms that jointly manage staleness and workload balance.

\noindent\textbf{Analogy to Asynchronous RL}. 
These challenges are related to synchronous LLM RL systems~\citep{sheng2025hybridflow,nemo-rl,hu2026jigsawrlassemblingrlpipelines}, which also suffer from synchronization overhead and workload imbalance. Asynchronous RL addresses this by overlapping rollout generation with training and controlling off-policy optimization~\citep{fu2025areal,zhong2025streamrl,sheng2025laminar}. Agent evolution differs in two key ways. First, it contains multiple LLM inference stages rather than a single "rollout" stage in RL. Each stage has batched generation behavior and its own long-tail imbalance. Second, staleness occurs over inspectable language artifacts, such as prompts, memories, harness code, and programs, rather than continuous model weights. This allows a more flexible design space for staleness handling policies.

\section{\tool{}: Asynchronous Framework for Agent Evolution}

\begin{figure}
    \centering
    \includegraphics[width=\linewidth]{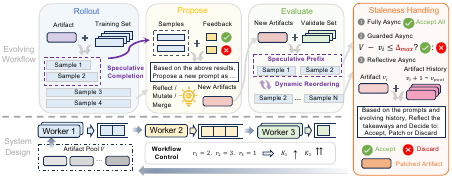}
    \caption{Overview of \tool{}. \tool{} executes agent evolution with asynchronous
workers and queues across stages. Workers pass partial or completed results through queues instead of waiting for a whole stage to finish. \tool{} further uses speculative completion, validation-set
reordering, workflow control, and staleness-aware handling to improve throughput while limiting data staleness.}
    \label{fig:placeholder}
\end{figure}

We present \tool{}, an asynchronous framework that removes the sequential and imbalanced behavior identified in Section~\ref{sec:inefficiency}. \tool{} decomposes an evolution loop into asynchronous workers connected by queues, so different stages and evolution steps can overlap. Each queue item carries the artifact state and pool version, allowing \tool{} to detect stale items. On top of this execution model, \tool{} introduces staleness-aware data handling, speculative stage completion, and adaptive workflow control to improve the time efficiency of evolution.

\subsection{Asynchronous Execution with Workers and Queues}
\label{sec:async}

\noindent\textbf{Asynchronous workers.} \tool{} turns a synchronized evolution step into asynchronous workers connected by queues. Instead of waiting for artifact proposal, validation, and pool update to finish before starting the next step, workers continuously process ready items and pass their outputs to downstream queues. This allows different stages and evolution steps to overlap. Each stage has an input queue and a set of workers. A queue item carries the artifact being evolved, the input/output, and the artifact-pool version $v_i$ at item creation. The pool version increases after each pool update, so \tool{} can compare $v_i$ with the current version $v$ to detect stale items.

\noindent\textbf{Worker concurrency.}
To improve system throughput, \tool{} assigns a worker count $K_i$ to each
asynchronous stage $i$. A larger $K_i$ allows more tasks in stage $i$ to issue
LLM requests at the same time, which increases per-stage concurrency so the whole
pipeline is not bottlenecked by the throughput of a single slow or imbalanced
stage. The tradeoff is data staleness: larger worker counts increase the chance that queued items were generated from an older artifact pool state.

\subsection{Staleness-Aware Data Handling}

\label{sec:staleness}

\tool{} supports three policies for handling such stale items with different tradeoffs:

\begin{itemize}[leftmargin=*]
    \item \textit{Full Async} does not check artifact pool versions and allows all
    items to continue through the pipeline. This policy preserves all completed work
    and maximizes throughput, but stale items may introduce outdated updates into
    the artifact pool and impact convergence.

    \item \textit{Guarded Async} discards an item when its version gap exceeds
    a threshold $\Delta_{\max}$. Let $v_i$ denote the artifact-pool version used
    to generate item $i$, and let $v$ denote the current artifact-pool version.
    The version gap is defined as $\Delta_i = v - v_i$. Guarded Async allows
    item $i$ to continue only when $\Delta_i \le \Delta_{\max}$; otherwise, it
    discards the item. This policy prevents highly stale items, but will waste the generated tokens that already spent on discarded items.

    \item \textit{Reflective Async} inspects and updates stale items by adding a
    new reflection worker stage. For an item $i$ with version gap
    $\Delta_i > 0$, the reflection worker uses the stale item and all
    artifact-pool updates between versions $v_i$ and $v$ to decide whether the
    item still contributes a useful change. If so, it patches the item against
    the current artifact pool state and lets it continue; otherwise, \tool{}
    discards it. Non-stale items continue without reflection. This policy avoids uncontrolled stale updates while reusing useful stale items, reducing wasted LLM generations.
\end{itemize}

\paragraph{Why language-space staleness can be repaired.}
Language-space staleness is discrete and inspectable, unlike parameter staleness in asynchronous RL, which is continuous and opaque. In RL, a stale item is tied to an older point in weight space, so systems typically handle it through importance weighting, bounded delay, or discard. In agent evolution, a stale item is text or code, such as a prompt edit, memory update, harness mutation, or generated program. \tool{} can therefore inspect the stale item together with the intervening artifact history and decide whether the edit is orthogonal, already subsumed, or conflicting with the current artifact pool. This makes repair a first-class operation: stale items can be patched when they contain reusable information, or discarded when they are too specific or inconsistent. Figure~\ref{fig:staleness} shows an example where \tool{} filters task-specific stale content and keeps transferable principles to form a compact prompt patch.

\subsection{Speculative Stage Completion}
\label{sec:speculative}

Asynchronous workers remove waiting between stages, but each worker may still wait for all LLM requests in its current stage before writing to the next queue. This still creates an intra-stage synchronization barrier, especially in rollout and evaluate stages where a minibatch contains many LLM requests. To reduce this barrier, \tool{} allows a stage to release partial output after a fraction $\alpha_{\mathrm{spec}}^{i} \in (0,1]$ of its requests has finished. The worker packages the completed results as a tentative queue item and continues the remaining requests in the background, while downstream workers can start from the tentative item.

For rollout, this means completed samples can be forwarded as soon as they are available. For evaluation, \tool{} adds a score threshold to avoid forwarding weak candidates. After the first $\alpha_{\mathrm{spec}}^{i}$ fraction of evaluation requests finishes, the worker computes a partial score. If the partial score exceeds the current pool score, \tool{} inserts the candidate into the pool as a \textit{speculative artifact}. When full evaluation finishes, the artifact is confirmed if it still passes the acceptance condition; otherwise, it is removed. If a speculative artifact is later removed, downstream items derived from it are marked stale and handled by the same staleness-aware policy in Section~\ref{sec:staleness}; they cannot update the confirmed pool without passing the normal validation path.

\noindent\textbf{Validation-set reordering.}
Speculative completion is more reliable when the early validation samples are informative. We call the first $\alpha_{\mathrm{spec}}$ fraction of the validation set the \emph{speculative prefix}. \tool{} reorders the validation set using sample pass history: samples that pass for $w$ consecutive rounds are moved out of the speculative prefix and placed later in the validation order. This keeps easy samples from dominating the early signal and leaves more discriminative samples in the prefix. We set $w=3$ to avoid reacting to one-round noise while keeping the prefix responsive to artifact improvement.

\subsection{Adaptive Workflow Control}
\label{sec:flowcontrol}

Different stages in an evolution loop produce and consume items at different
rates. A stage with short LLM requests can quickly fill the queue of a later
stage whose requests are longer or more imbalanced. If workers keep running at
a fixed concurrency, the queue keeps growing and many items become stale before
they are processed. \tool{} therefore monitors queue pressure and version gap to adjust worker behavior, making execution more
balanced and token efficient.

\noindent\textbf{Adaptive worker reallocation.}
\tool{} measures the item production rate of each asynchronous stage. The
production rate is the number of queue items that a stage writes to its
downstream queue per second. A stage with a much lower production rate can
limit the whole workflow, while a stage with a much higher production rate can
overfeed downstream queues.

\tool{} compares production rates across stages and adjusts their worker
counts. If a stage produces items at less than half
the median stage rate, we increase its worker count. If a stage produces
items at more than twice the median stage rate, we decrease its worker
count. Each adjustment changes the worker count by at most one, and each stage
has a minimum and maximum worker count. This avoids large swings while still
correcting persistent throughput imbalance.

\subsection{Implementation}
\label{sec:implementation}

\tool{} is implemented in Python with lightweight threads and in-process queues. Each stage is executed by a small worker pool, queue items carry the artifact state and pool version, and pool updates are applied under a lock. For a fair comparison, we run all open-source baselines and \tool{} on the same LLM serving stack: the native LLM calls in different algorithms are replaced by the same \texttt{DSPy} client backed by a local vLLM~\cite{kwon2023vllm} server with an OpenAI-compatible endpoint. Thus all methods benefit from the same continuous batching and KV-cache reuse, and throughput differences mainly reflect the optimization of the evolution pipeline. The same interface is also used for API-based experiments by changing only the endpoint and model name.

\section{Evaluation}

\begin{table*}[t]
\centering
\small
\caption{Throughput comparison on GEPA workloads. LLM throughput measures the output token rate of the whole system. Proposal throughput measures the rate of new candidate artifact generation.}
\label{tab:throughput}
\setlength{\tabcolsep}{4pt}
\begin{tabular}{l rrrr rrrr}
\toprule
& \multicolumn{4}{c}{LLM Throughput (token/s)}
& \multicolumn{4}{c}{Proposal Throughput (proposal/min)} \\
\cmidrule(lr){2-5}\cmidrule(lr){6-9}
Method & IFBench & HotpotQA & HoVer & AIME & IFBench & HotpotQA & HoVer & AIME \\
\midrule
\rowcolor{gray!12}
\multicolumn{9}{l}{\textit{vLLM with Qwen3-8B}} \\
GEPA    & 963 & 30 & 461 & 200 & 1.9 & 4.6 & 2.5 & 2.2 \\
Combee ($K{=}10$)          & 696 & 38 & 810 & 994 & 1.2 & 2.7 & 2.0 & 6.2 \\
Combee ($K{=}40$)          & 900 & 44 & 891 & 977 & 0.7 & 4.5 & 2.0 & 1.6 \\
\tool{}                    & \textbf{2{,}688} & \textbf{93} & \textbf{1{,}255} & \textbf{998}
                           & \textbf{8.9} & \textbf{8.8} & \textbf{5.9} & \textbf{11.4} \\
\midrule
\rowcolor{gray!12}
\multicolumn{9}{l}{\textit{API with GPT-4o-mini}} \\
GEPA  & 361 & 14 & 142 & 103 & 1.7 & 2.4 & 1.8 & 1.3 \\
Combee ($K{=}10$)            & 397 & 18 & 348 & 211 & 1.0 & 1.4 & 0.8 & 1.0 \\
Combee ($K{=}40$)          & 389 & 23 & 214 & 336 & 0.8 & 1.2 & 0.7 & 0.6 \\
\tool{}                    & \textbf{791} & \textbf{32} & \textbf{352} & \textbf{485}
                           & \textbf{10.1} & \textbf{8.0} & \textbf{9.1} & \textbf{6.6} \\
\bottomrule
\end{tabular}
\end{table*}

\subsection{Experimental Setup}

\noindent \textbf{Evolving algorithm baselines}. We evaluate \tool{} on three test-time evolution algorithms that optimize different artifacts. We use GEPA~\citep{agrawal2025gepa}, which evolves prompts through execution
feedback and reflection, as the main algorithm for end-to-end comparison and
ablation studies. We set the rollout minibatch size $mb$ = $3$, following the
default setting. We also reproduce Combee~\citep{li2026combee} as a scaling-oriented baseline.
Combee scales batch-level parallelism to improve throughput. We evaluate
its reported fixed-batch variants with parallelization sizes $B=$ $10$ and $40$.

We also evaluate \tool{} on ACE~\citep{zhang2025ace} and
Meta-Harness~\citep{lee2026metaharness}. ACE evolves agent context playbooks,
while Meta-Harness evolves harness code. These algorithms use different artifact types, but all fall into the abstraction optimized by \tool{}: a multi-stage evolution loop with LLM-heavy stages, queueable intermediate results, and a shared artifact pool
that is updated over time.

\noindent \textbf{Models and deployment}.
For open-source model experiments, we use Qwen3-8B~\cite{yang2025qwen3}, which is the default model
used in GEPA and provides a representative setting for studying test-time
evolution behavior. We serve Qwen3-8B with vLLM on a single NVIDIA H100 80GB
GPU and an AMD EPYC 9534 CPU. For API-based experiments, we use GPT-4o-mini~\cite{hurst2024gpt}, which shows similar evolution behavior to the open-source setting while representing a common remote-serving deployment.

\noindent \textbf{Datasets}. We use the benchmark datasets used in each original evolution algorithm. 
For GEPA, we evaluate on IFBench~\citep{pyatkin2025ifbench} for 
instruction following, HotpotQA~\citep{yang2018hotpotqa} for knowledge 
retrieval, HoVer~\citep{jiang2020hover} for multi-hop verification, 
and AIME for mathematical reasoning. For ACE, we use 
FiNER~\citep{loukas2022finer} and FormulaReasoning, 
which are finance and numerical-reasoning datasets, respectively. 
For Meta-Harness, we use a mixture of 
Symptom2Disease for medical diagnosis classification and 
AGNews~\citep{zhang2015character} for topic categorization. 
These datasets cover diverse domains of agent applications.

\subsection{Improvement on GEPA}
\label{sec:eval-gepa}

\noindent \textbf{System throughput improvement}. Table~\ref{tab:throughput} first shows that \tool{} substantially improves LLM
throughput. This indicates that asynchronous stage orchestration keeps the LLM
backend busier by overlapping requests from different stages and evolution
steps. On local vLLM serving, \tool{} improves LLM throughput by
$3.4\times$ on average over GEPA and $1.9\times$ on average over the best
Combee setting. The same trend holds for API-based serving: \tool{} improves
LLM throughput by $2.9\times$ on average over GEPA and $1.5\times$ on average
over the best Combee setting.

We also show that higher LLM throughput translates
into faster artifact exploration. On local vLLM serving, \tool{} improves
proposal throughput by $3.5\times$ on average over GEPA and $3.5\times$ on
average over the best Combee setting. On API-based serving, \tool{} improves
proposal throughput by $4.9\times$ on average over GEPA and $8.4\times$ on
average over the best Combee setting. Across all settings, \tool{} sustains
more than 5.9 proposals/min and up to 11.4 proposals/min, showing that it
substantially increases the rate at which evolution tests new candidate
artifacts.

\begin{figure}[t]
    \centering
    \includegraphics[width=\linewidth]{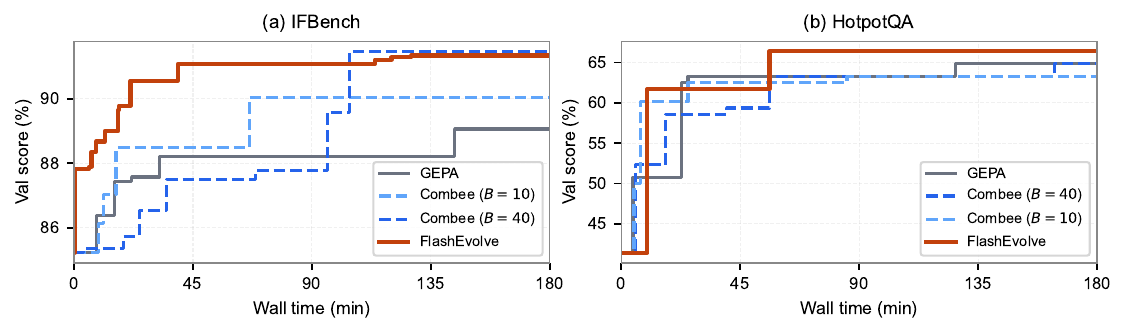}
    \caption{Longer-time validation score evolution over wall-clock time with Qwen3-8B.}
    \label{fig:long-term}
\end{figure}

\noindent \textbf{Evolving efficiency improvement}. 
Table~\ref{tab:main-results} reports validation score and normalized evolution rate within a fixed 30-minute budget. Across the three workloads where the GEPA baseline makes measurable progress, \tool{} achieves an average normalized evolution rate of $1.43\times$. The strongest gain appears on IFBench, where \tool{} improves the validation score from 87.6 to 90.6 and reaches a $2.27\times$ normalized evolution rate. On HoVer, \tool{} also achieves the best score and a $1.15\times$ normalized rate. On HotpotQA, \tool{} does not reach the best validation score within the 30-minute budget, but Figure~\ref{fig:long-term} shows that this advantage emerges under a longer budget. On AIME, both GEPA and Combee remain at the initial score of 10.0\%, while \tool{} reaches 15.0\%, making it the only method that improves over the initial score.

\begin{table*}[t]
\centering
\small
\caption{Validation score(\%) and normalized evolution rate within 30 minutes on GEPA workloads using Qwen3-8B. Evolution rate measures the score improvement achieved within the time budget, normalized by the improvement achieved by serial GEPA, reflecting the speedup of evolution progress. AIME reports ``--'' because serial GEPA shows no score improvement within the budget.}
\label{tab:main-results}
\setlength{\tabcolsep}{4pt}
\begin{tabular}{l cc cc cc cc}
\toprule
& \multicolumn{2}{c}{IFBench}
& \multicolumn{2}{c}{HoVer}
& \multicolumn{2}{c}{HotpotQA}
& \multicolumn{2}{c}{AIME} \\
\cmidrule(lr){2-3}
\cmidrule(lr){4-5}
\cmidrule(lr){6-7}
\cmidrule(lr){8-9}
Method & Score & Norm.\ Rate & Score & Norm.\ Rate & Score & Norm.\ Rate & Score & Norm.\ Rate \\
\midrule
GEPA
& 87.6 & 1.00
& 39.8 & 1.00
& \textbf{63.3} & \textbf{1.00}
& 10.0 & -- \\

Combee ($K{=}10$)
& 88.5 & 1.39
& 41.2 & 1.09
& 62.5 & 0.94
& 10.0 & -- \\

Combee ($K{=}40$)
& 86.5 & 0.55
& 40.5 & 1.05
& 58.6 & 0.63
& 10.0 & -- \\

\tool{}
& \textbf{90.6} & \textbf{2.27}
& \textbf{42.0} & \textbf{1.15}
& 61.7 & 0.88
& \textbf{15.0} & -- \\
\bottomrule
\end{tabular}
\end{table*}

\noindent \textbf{Long-time evolution}. 
Figure~\ref{fig:long-term} reports validation score over a longer 180-minute budget. \tool{} reaches strong validation scores earlier than the synchronous baselines. On IFBench, \tool{} reaches 91\% in 39.3 minutes, while Combee ($B{=}40$) reaches the same score region after 104.2 minutes and eventually approaches a similar final score. On HotpotQA, \tool{} reaches its best score of 66.41\% at 56.1 minutes and maintains the highest validation score over the full budget, while all baselines remain below 65\%. These results show that asynchronous evolution accelerates useful artifact discovery, and in some workloads also improves the final score under a longer time budget.

\subsection{Ablation Studies}
\label{sec:eval-ablation}

\begin{figure}[t]
    \centering
    \includegraphics[width=\linewidth]{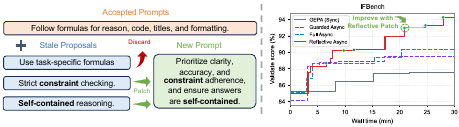}
    \caption{Staleness handling on IFBench with Qwen3-8B. The left panel shows an example of reflective prompt repair. The right panel compares Full Async, Guarded Async, and Reflective Async.}
    \label{fig:staleness}
\end{figure}

\noindent \textbf{Comparison across staleness handling methods}.
\label{sec:eval-staleness}
Figure~\ref{fig:staleness} compares three staleness handling policies on IFBench. Full Async and Guarded Async achieve similar final scores, but their behaviors differ. Full Async preserves all stale items and therefore keeps high throughput, while Guarded Async discards highly stale items to avoid outdated updates. In this case, the two methods perform similarly.

Reflective Async achieves the best evolution efficiency. It reaches a validation score of 94.3\% within the 30-minute budget. This shows that stale items are not always useless: when the stale artifact is text, \tool{} can inspect it, discard task-specific noise, and reuse transferable principles. Figure~\ref{fig:staleness} also shows a simplified repair example extracted from our logs. \tool{} discards task-specific formulas because the accepted prompt already contains general instruction-following rules and the formula does not transfer across tasks. In contrast, new takeaways such as stricter constraint checking and self-contained reasoning are distilled into a compact prompt patch. We also observe that many score jumps in the Reflective Async curve come from prompts after such patches, suggesting that reflective repair improves the quality of prompt proposals rather than only increasing throughput.

\begin{figure}[t]
    \centering
    \includegraphics[width=\linewidth]{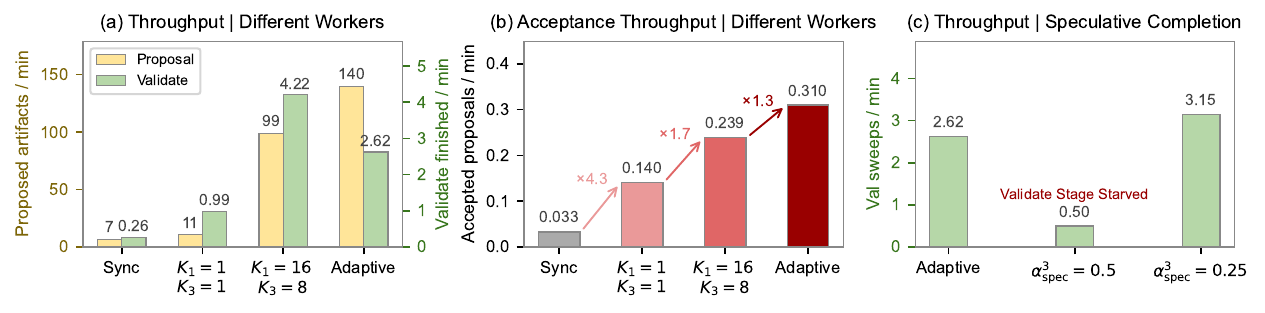}
    \caption{Ablation of worker concurrency and speculative completion on IFBench with Qwen3-8B. 
    (a) Worker allocation varies throughput. (b) Adaptive worker control achieves the highest accepted proposal throughput by balancing proposal generation and validation. 
    (c) Speculative completion improves validation throughput when the prefix threshold is properly set.}
    \label{fig:ablation}
\end{figure}

\noindent \textbf{Concurrent Workers}.
We study how worker counts affect stage throughput. As shown in Figure~\ref{fig:ablation}(a), larger worker counts greatly increase proposal throughput, from 7 artifacts/min in the synchronous setting to 99 artifacts/min with $K_1{=}16,K_3{=}8$. However, validation throughput does not scale uniformly, showing that naive scaling can shift the bottleneck across stages. Adaptive control balances queue pressure and stage rates; although it reduces validation throughput compared with the fixed large-worker setting, it achieves the highest accepted proposal throughput in Figure~\ref{fig:ablation}(b), suggesting that a more balanced worker allocation can produce a higher rate of high-quality candidates rather than merely increasing raw proposal volume.

\noindent \textbf{Speculative Stage Completion}.
Figure~\ref{fig:ablation}(c) shows that speculative completion can improve validation throughput when the prefix threshold is properly set. With $\alpha_{\mathrm{spec}}^3{=}0.25$, validation-stage throughput increases to 3.15 validations/min and the validation score improves by 4.49 percentage points within 30 minutes. However, when $\alpha_{\mathrm{spec}}^3{=}0.5$, the speculative gate becomes less effective: candidates must wait for a larger partial validation prefix before being released, which reduces early handoff and lowers validation throughput. Overall, speculative completion can be useful in some settings, but its accuracy and efficiency depend on $\alpha_{\mathrm{spec}}$ and dataset characteristics. We therefore treat it as an optional optimization in \tool{} rather than include it in the main evaluation.

\noindent \textbf{}

\subsection{Improvement on Other Algorithms: ACE and Meta-Harness}
\label{sec:eval-generality}

\tool{} is algorithm-agnostic. It does not rely on a specific artifact type, but only assumes that the evolution loop contains multiple stages that need orchestration. We evaluate \tool{} on two other algorithms including ACE~\citep{zhang2025ace} and Meta-Harness~\citep{lee2026metaharness}, which evolves harness code.

\begin{figure}[t]
    \centering
    \includegraphics[width=\linewidth]{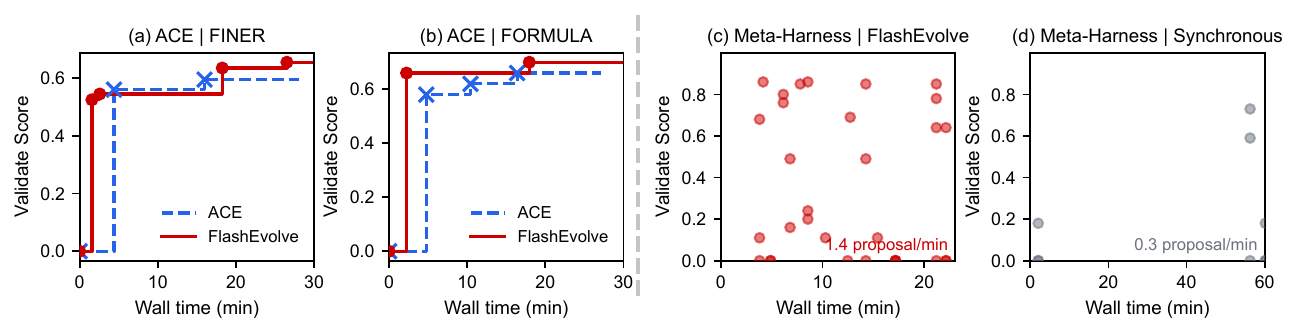}
    \caption{\tool{} on other algorithms for agent evolution. (a) and (b) compare validation score evolution curve on FiNER and Formula over a fixed time budget. (c) and (d) compare Meta-Harness proposal rate and validation scores of different proposals on Symptom2Disease and AgNews. }
    \label{fig:generality}
\end{figure}

Figure~\ref{fig:generality}(a)(b) shows the wall-clock evolution curves of ACE. \tool{} reaches better validation scores within the same 30-minute budget on both tasks. The validation score improves from $0.6$ to $0.66$ on FiNER and from $0.66$ to $0.7$ on Formula, demonstrating higher efficiency.

Figure~\ref{fig:generality}(c)(d) compares \tool{} with the synchronous Meta-Harness baseline. \tool{} improves the proposal and validation throughput from $0.3$ to $1.4$ proposals/min, giving a $4.7\times$ speedup. Since the open-source model has relatively weak code-generation capability, harness-code evolution progresses slowly in both settings. We therefore report the score distribution of different proposed harnesses. With higher proposal throughput, \tool{} samples and validates more harness candidates within the same time budget, leading to a higher potential of improvement.

\section{Conclusion}

This paper presents \tool{}, a framework for accelerating agent evolution in wall-clock time. \tool{} replaces synchronous stage execution with asynchronous workers and queues, allowing LLM-heavy stages and evolution steps to overlap. It preserves evolution semantics through artifact-pool versioning and staleness-aware policies that update, discard, or patch stale artifacts, and further improves efficiency with speculative stage completion and adaptive workflow control. On GEPA workloads, \tool{} achieves $3.5\times$ higher proposal throughput over the synchronous implementation on local vLLM serving. The same execution model also generalizes to context evolution with ACE and harness-code evolution with Meta-Harness,

\noindent \textbf{Limitations}. 
\tool{} currently supports only a limited set of agent-evolution algorithms, and each integration still requires algorithm-specific implementation effort. Although the worker and queue abstraction is general, mapping a new algorithm to this abstraction requires implementing its stages, queue items, artifact state, and update rules. Our current evaluation also focuses on representative prompt, context, and harness-code evolution workloads, and broader coverage of evolution algorithms and artifact types remains future work.

\noindent \textbf{Future Work}. 
In future work, we plan to expand \tool{} with a more general plugin interface for defining stages, artifacts, staleness policies, and evaluation logic, so that new evolution algorithms can be integrated with less manual engineering. We also plan to extend \tool{} to more types of artifact evolution, such as memory, tool-use policies and generated programs.


\bibliographystyle{abbrvnat}
\bibliography{ref}

@article{agrawal2025gepa,
  title={Gepa: Reflective prompt evolution can outperform reinforcement learning},
  author={Agrawal, Lakshya A and Tan, Shangyin and Soylu, Dilara and Ziems, Noah and Khare, Rishi and Opsahl-Ong, Krista and Singhvi, Arnav and Shandilya, Herumb and Ryan, Michael J and Jiang, Meng and others},
  journal={arXiv preprint arXiv:2507.19457},
  year={2025}
}

@article{zhang2025ace,
  title={Agentic context engineering: Evolving contexts for self-improving language models},
  author={Zhang, Qizheng and Hu, Changran and Upasani, Shubhangi and Ma, Boyuan and Hong, Fenglu and Kamanuru, Vamsidhar and Rainton, Jay and Wu, Chen and Ji, Mengmeng and Li, Hanchen and others},
  journal={arXiv preprint arXiv:2510.04618},
  year={2025}
}

@article{lee2026metaharness,
  title={Meta-Harness: End-to-End Optimization of Model Harnesses},
  author={Lee, Yoonho and Nair, Roshen and Zhang, Qizheng and Lee, Kangwook and Khattab, Omar and Finn, Chelsea},
  journal={arXiv preprint arXiv:2603.28052},
  year={2026}
}

@article{novikov2025alphaevolve,
  title={Alphaevolve: A coding agent for scientific and algorithmic discovery},
  author={Novikov, Alexander and V{\~u}, Ng{\^a}n and Eisenberger, Marvin and Dupont, Emilien and Huang, Po-Sen and Wagner, Adam Zsolt and Shirobokov, Sergey and Kozlovskii, Borislav and Ruiz, Francisco JR and Mehrabian, Abbas and others},
  journal={arXiv preprint arXiv:2506.13131},
  year={2025}
}

@article{ouyang2025reasoningbank,
  title={Reasoningbank: Scaling agent self-evolving with reasoning memory},
  author={Ouyang, Siru and Yan, Jun and Hsu, I and Chen, Yanfei and Jiang, Ke and Wang, Zifeng and Han, Rujun and Le, Long T and Daruki, Samira and Tang, Xiangru and others},
  journal={arXiv preprint arXiv:2509.25140},
  year={2025}
}

@article{zhang2025memevolve,
  title={Memevolve: Meta-evolution of agent memory systems},
  author={Zhang, Guibin and Ren, Haotian and Zhan, Chong and Zhou, Zhenhong and Wang, Junhao and Zhu, He and Zhou, Wangchunshu and Yan, Shuicheng},
  journal={arXiv preprint arXiv:2512.18746},
  year={2025}
}

@article{lange2025shinkaevolve,
  title={Shinkaevolve: Towards open-ended and sample-efficient program evolution},
  author={Lange, Robert Tjarko and Imajuku, Yuki and Cetin, Edoardo},
  journal={arXiv preprint arXiv:2509.19349},
  year={2025}
}

@article{assumpccao2025codeevolve,
  title={Codeevolve: An open source evolutionary coding agent for algorithm discovery and optimization},
  author={Assump{\c{c}}{\~a}o, Henrique and Ferreira, Diego and Campos, Leandro and Murai, Fabricio},
  journal={arXiv preprint arXiv:2510.14150},
  year={2025}
}

@article{wang2023promptagent,
  title={Promptagent: Strategic planning with language models enables expert-level prompt optimization},
  author={Wang, Xinyuan and Li, Chenxi and Wang, Zhen and Bai, Fan and Luo, Haotian and Zhang, Jiayou and Jojic, Nebojsa and Xing, Eric P and Hu, Zhiting},
  journal={arXiv preprint arXiv:2310.16427},
  year={2023}
}

@article{lou2026autoharness,
  title={AutoHarness: improving LLM agents by automatically synthesizing a code harness},
  author={Lou, Xinghua and L{\'a}zaro-Gredilla, Miguel and Dedieu, Antoine and Wendelken, Carter and Lehrach, Wolfgang and Murphy, Kevin P},
  journal={arXiv preprint arXiv:2603.03329},
  year={2026}
}

@article{gao2025survey,
  title={A survey of self-evolving agents: On path to artificial super intelligence},
  author={Gao, Huan-ang and Geng, Jiayi and Hua, Wenyue and Hu, Mengkang and Juan, Xinzhe and Liu, Hongzhang and Liu, Shilong and Qiu, Jiahao and Qi, Xuan and Wu, Yiran and others},
  journal={arXiv preprint arXiv:2507.21046},
  volume={1},
  year={2025}
}

@article{guo2025deepseekr1,
  title={Deepseek-r1: Incentivizing reasoning capability in llms via reinforcement learning},
  author={Guo, Daya and Yang, Dejian and Zhang, Haowei and Song, Junxiao and Wang, Peiyi and Zhu, Qihao and Xu, Runxin and Zhang, Ruoyu and Ma, Shirong and Bi, Xiao and others},
  journal={arXiv preprint arXiv:2501.12948},
  year={2025}
}

@article{xiao2025promptmii,
  title={Prompt-MII: Meta-Learning Instruction Induction for LLMs},
  author={Xiao, Emily and Zeng, Yixiao and Chen, Ada and Li, Chin-Jou and Bertsch, Amanda and Neubig, Graham},
  journal={arXiv preprint arXiv:2510.16932},
  year={2025}
}

@article{fang2025comprehensive,
  title={A comprehensive survey of self-evolving ai agents: A new paradigm bridging foundation models and lifelong agentic systems},
  author={Fang, Jinyuan and Peng, Yanwen and Zhang, Xi and Wang, Yingxu and Yi, Xinhao and Zhang, Guibin and Xu, Yi and Wu, Bin and Liu, Siwei and Li, Zihao and others},
  journal={arXiv preprint arXiv:2508.07407},
  year={2025}
}

@article{jaech2024openaio1,
  title={Openai o1 system card},
  author={Jaech, Aaron and Kalai, Adam and Lerer, Adam and Richardson, Adam and El-Kishky, Ahmed and Low, Aiden and Helyar, Alec and Madry, Aleksander and Beutel, Alex and Carney, Alex and others},
  journal={arXiv preprint arXiv:2412.16720},
  year={2024}
}

@article{shinn2023reflexion,
  title={Reflexion: Language agents with verbal reinforcement learning},
  author={Shinn, Noah and Cassano, Federico and Gopinath, Ashwin and Narasimhan, Karthik and Yao, Shunyu},
  journal={Advances in neural information processing systems},
  volume={36},
  pages={8634--8652},
  year={2023}
}

@article{madaan2023selfrefine,
  title={Self-refine: Iterative refinement with self-feedback},
  author={Madaan, Aman and Tandon, Niket and Gupta, Prakhar and Hallinan, Skyler and Gao, Luyu and Wiegreffe, Sarah and Alon, Uri and Dziri, Nouha and Prabhumoye, Shrimai and Yang, Yiming and others},
  journal={Advances in neural information processing systems},
  volume={36},
  pages={46534--46594},
  year={2023}
}

@article{ouyang2022rlhf,
  title={Training language models to follow instructions with human feedback},
  author={Ouyang, Long and Wu, Jeffrey and Jiang, Xu and Almeida, Diogo and Wainwright, Carroll and Mishkin, Pamela and Zhang, Chong and Agarwal, Sandhini and Slama, Katarina and Ray, Alex and others},
  journal={Advances in neural information processing systems},
  volume={35},
  pages={27730--27744},
  year={2022}
}

@article{shoeybi2019megatron,
  title={Megatron-lm: Training multi-billion parameter language models using model parallelism},
  author={Shoeybi, Mohammad and Patwary, Mostofa and Puri, Raul and LeGresley, Patrick and Casper, Jared and Catanzaro, Bryan},
  journal={arXiv preprint arXiv:1909.08053},
  year={2019}
}

@article{zhao2023pytorchfsdp,
  title={Pytorch fsdp: experiences on scaling fully sharded data parallel},
  author={Zhao, Yanli and Gu, Andrew and Varma, Rohan and Luo, Liang and Huang, Chien-Chin and Xu, Min and Wright, Less and Shojanazeri, Hamid and Ott, Myle and Shleifer, Sam and others},
  journal={arXiv preprint arXiv:2304.11277},
  year={2023}
}

@inproceedings{kwon2023vllm,
  title={Efficient memory management for large language model serving with pagedattention},
  author={Kwon, Woosuk and Li, Zhuohan and Zhuang, Siyuan and Sheng, Ying and Zheng, Lianmin and Yu, Cody Hao and Gonzalez, Joseph and Zhang, Hao and Stoica, Ion},
  booktitle={Proceedings of the 29th symposium on operating systems principles},
  pages={611--626},
  year={2023}
}

@article{pyatkin2025ifbench,
  title={Generalizing verifiable instruction following},
  author={Pyatkin, Valentina and Malik, Saumya and Graf, Victoria and Ivison, Hamish and Huang, Shengyi and Dasigi, Pradeep and Lambert, Nathan and Hajishirzi, Hannaneh},
  journal={arXiv preprint arXiv:2507.02833},
  year={2025}
}

@inproceedings{yang2018hotpotqa,
  title={HotpotQA: A dataset for diverse, explainable multi-hop question answering},
  author={Yang, Zhilin and Qi, Peng and Zhang, Saizheng and Bengio, Yoshua and Cohen, William and Salakhutdinov, Ruslan and Manning, Christopher D},
  booktitle={Proceedings of the 2018 conference on empirical methods in natural language processing},
  pages={2369--2380},
  year={2018}
}

@misc{hu2026jigsawrlassemblingrlpipelines,
      title={JigsawRL: Assembling RL Pipelines for Efficient LLM Post-Training}, 
      author={Zhengding Hu and Hehua Ouyang and Chang Chen and Zaifeng Pan and Yue Guan and Zhongkai Yu and Zhen Wang and Steven Swanson and Yufei Ding},
      year={2026},
      eprint={2604.23838},
      archivePrefix={arXiv},
      primaryClass={cs.LG},
      url={https://arxiv.org/abs/2604.23838}, 
}

@inproceedings{sheng2025hybridflow,
  title={Hybridflow: A flexible and efficient rlhf framework},
  author={Sheng, Guangming and Zhang, Chi and Ye, Zilingfeng and Wu, Xibin and Zhang, Wang and Zhang, Ru and Peng, Yanghua and Lin, Haibin and Wu, Chuan},
  booktitle={Proceedings of the Twentieth European Conference on Computer Systems},
  pages={1279--1297},
  year={2025}
}

@misc{nemo-rl,
title = {NeMo RL: A Scalable and Efficient Post-Training Library},
howpublished = {\url{https://github.com/NVIDIA-NeMo/RL}},
year = {2025},
note = {GitHub repository},
}

@article{fu2025areal,
  title={Areal: A large-scale asynchronous reinforcement learning system for language reasoning},
  author={Fu, Wei and Gao, Jiaxuan and Shen, Xujie and Zhu, Chen and Mei, Zhiyu and He, Chuyi and Xu, Shusheng and Wei, Guo and Mei, Jun and Wang, Jiashu and others},
  journal={arXiv preprint arXiv:2505.24298},
  year={2025}
}

@article{zhong2025streamrl,
  title={Streamrl: Scalable, heterogeneous, and elastic rl for llms with disaggregated stream generation},
  author={Zhong, Yinmin and Zhang, Zili and Song, Xiaoniu and Hu, Hanpeng and Jin, Chao and Wu, Bingyang and Chen, Nuo and Chen, Yukun and Zhou, Yu and Wan, Changyi and others},
  journal={arXiv preprint arXiv:2504.15930},
  year={2025}
}

@article{sheng2025laminar,
  title={Laminar: A scalable asynchronous rl post-training framework},
  author={Sheng, Guangming and Tong, Yuxuan and Wan, Borui and Zhang, Wang and Jia, Chaobo and Wu, Xibin and Wu, Yuqi and Li, Xiang and Zhang, Chi and Peng, Yanghua and others},
  journal={arXiv preprint arXiv:2510.12633},
  year={2025}
}

@article{hurst2024gpt,
  title={Gpt-4o system card},
  author={Hurst, Aaron and Lerer, Adam and Goucher, Adam P and Perelman, Adam and Ramesh, Aditya and Clark, Aidan and Ostrow, AJ and Welihinda, Akila and Hayes, Alan and Radford, Alec and others},
  journal={arXiv preprint arXiv:2410.21276},
  year={2024}
}

@article{yang2025qwen3,
  title={Qwen3 technical report},
  author={Yang, An and Li, Anfeng and Yang, Baosong and Zhang, Beichen and Hui, Binyuan and Zheng, Bo and Yu, Bowen and Gao, Chang and Huang, Chengen and Lv, Chenxu and others},
  journal={arXiv preprint arXiv:2505.09388},
  year={2025}
}

@article{li2026combee,
  title={Combee: Scaling Prompt Learning for Self-Improving Language Model Agents},
  author={Li, Hanchen and He, Runyuan and Zhang, Qizheng and Ji, Changxiu and Mang, Qiuyang and Chen, Xiaokun and Agrawal, Lakshya A and Liao, Wei-Liang and Yang, Eric and Cheung, Alvin and others},
  journal={arXiv preprint arXiv:2604.04247},
  year={2026}
}

@article{zheng2024sglang,
  title={Sglang: Efficient execution of structured language model programs},
  author={Zheng, Lianmin and Yin, Liangsheng and Xie, Zhiqiang and Sun, Chuyue and Huang, Jeff and Yu, Cody H and Cao, Shiyi and Kozyrakis, Christos and Stoica, Ion and Gonzalez, Joseph E and others},
  journal={Advances in neural information processing systems},
  volume={37},
  pages={62557--62583},
  year={2024}
}

@article{lu2026empirical,
  title={Empirical-MCTS: Continuous Agent Evolution via Dual-Experience Monte Carlo Tree Search},
  author={Lu, Hao and Huang, Haoyuan and Zhou, Yulin and Li, Chen and Zhu, Ningxin},
  journal={arXiv preprint arXiv:2602.04248},
  year={2026}
}

@article{yuksekgonul2024textgrad,
  title={Textgrad: Automatic" differentiation" via text},
  author={Yuksekgonul, Mert and Bianchi, Federico and Boen, Joseph and Liu, Sheng and Huang, Zhi and Guestrin, Carlos and Zou, James},
  journal={arXiv preprint arXiv:2406.07496},
  year={2024}
}

@inproceedings{jiang2020hover,
  title={HoVer: A dataset for many-hop fact extraction and claim verification},
  author={Jiang, Yichen and Bordia, Shikha and Zhong, Zheng and Dognin, Charles and Singh, Maneesh and Bansal, Mohit},
  booktitle={Findings of the Association for Computational Linguistics: EMNLP 2020},
  pages={3441--3460},
  year={2020}
}

@inproceedings{loukas2022finer,
  title={FiNER: Financial numeric entity recognition for XBRL tagging},
  author={Loukas, Lefteris and Fergadiotis, Manos and Chalkidis, Ilias and Spyropoulou, Eirini and Malakasiotis, Prodromos and Androutsopoulos, Ion and Paliouras, Georgios},
  booktitle={Proceedings of the 60th Annual Meeting of the Association for Computational Linguistics (Volume 1: Long Papers)},
  pages={4419--4431},
  year={2022}
}

@article{zhang2015character,
  title={Character-level convolutional networks for text classification},
  author={Zhang, Xiang and Zhao, Junbo and LeCun, Yann},
  journal={Advances in neural information processing systems},
  volume={28},
  year={2015}
}


\appendix





\end{document}